\documentclass[]{interact}

\usepackage{epstopdf}
\usepackage[caption=false]{subfig}

\usepackage[natbibapa,nodoi]{apacite}
\theoremstyle{plain}

\theoremstyle{definition}

\theoremstyle{remark}

\usepackage[hyphens]{url}
\usepackage[
	colorlinks=true,
	linkcolor=blue,
	citecolor=blue,
	urlcolor=blue]{hyperref}
	
\usepackage{multirow}

\usepackage{soul}

\usepackage{amssymb}

\begin{document}

\articletype{}


\title{An algorithm for the selection of route dependent orientation information}

\author{
\name{Heinrich L{\"o}wen\textsuperscript{a}\thanks{CONTACT Heinrich L{\"o}wen. Email: loewen.heinrich@uni-muenster.de} and Angela Schwering\textsuperscript{a}}
\affil{\textsuperscript{a}Institute for Geoinformatics, University of M{\"u}nster, Germany}
}

\maketitle

\begin{abstract}
Landmarks are important features of spatial cognition. Landmarks are naturally included in human route descriptions and in the past algorithms were developed to select the most salient landmarks at decision points and automatically incorporate them in route instructions. 
Moreover, it was shown that human route descriptions contain a significant amount of orientation information and that these orientation information support the acquisition of survey knowledge. 
Thus, there is a need to extend the landmarks selection in order to automatically select orientation information.
In this work we present an algorithm for the computational selection of route dependent orientation information, which extends previous algorithms and includes a salience evaluation of orientation information for any location along the route.
We implemented the algorithm and demonstrate the functionality on the basis of OpenStreetMap data. 
\end{abstract}


\section{Introduction}

When asked for a route description, humans naturally refer to landmarks and orientation information in order to help the navigator to orient themselves and find their way \citep{Anacta2016}.
In contrast, wayfinding support systems rarely incorporate orientation information, but rather rely on metric information which is communicated to the navigator step-by-step at the required locations along the route.
While the broad availability of these systems fundamentally changed peoples' wayfinding behaviour, it was shown that this has negative consequences on human spatial abilities \citep{Munzer2012,Kruger2004,Ishikawa2008,Burnett2005}.
People tend to blindly follow the instructions of the navigation devices without active engagement with the environment, which makes them unable to orient themselves even after recurrent drives. 
However, recent research has shown that the accentuation of orientation information has a significant effect on peoples' spatial knowledge acquisition during wayfinding.
When people are presented with global and structuring information during navigation, these information are automatically learned, which leads a positive effect on peoples' survey knowledge \citep{Lowen2019}.
 
Previous research proposed computational methods to identify the most salient landmarks at decision points and presented approaches to automatically incorporate these landmarks in routing instructions. 
However, the proposed solutions ignore the importance of context and are restricted to the selection of point-like landmarks at decision points.
There is the need for an algorithm to automatically select the most salient environmental features for any location along a route also considering the task to be supported, e.g. conveying orientation in the local or global context.
Orientation information, which includes local and global landmarks and even structuring features like environmental regions, supports people to orient themselves in the current context.
In this paper, we present an algorithm to select orientation information candidates from spatial databases and evaluate the salience of these candidates with respect to the current route context (Section~\ref{sec:algorithm}). 
We discuss, how the algorithm can be modified and implemented into navigation support systems.
As a proof of concept, we implement the algorithm and demonstrate the selection and salience evaluation based on OpenStreetMap (OSM) data (Section~\ref{sec:example}).
We review and discuss the results in the light of the extendability and generalizability of the algorithm (Section~\ref{sec:results} and \ref{sec:discussion}).

\section{Related Works}
\label{sec:related_works}

\subsection{Algorithms for the Salience Evaluation of Landmarks}

In the past, several methods were developed to investigate the salience of landmarks and to automatically include the most salient landmarks in wayfinding instructions.
Existing algorithms, however, only consider the selection of point-like landmarks at decision points. 
\citet{Sorrows1999} reviewed the nature of landmarks and proposed to distinguish \textit{visual}, \textit{cognitive}, and \textit{structural} landmarks, which affect users' navigation in space in different ways. Moreover, they discussed that these categories are not discrete, but landmarks can have properties of all three categories and the best landmarks are the ones that are prominent in all three categories. 
Building upon this, \citet{Raubal2002} proposed measures to formally assess the salience of landmarks at decision points. 
They defined measures for \textit{visual}, \textit{semantic}, and \textit{structural salience} in order to assess the overall salience of a landmark. 
For the visual salience, they considered the \textit{facade area}, the \textit{shape}, the \textit{color}, and the \textit{visibility} of objects and calculate the prominence of the objects in terms of these metrics.
For the semantic salience, they evaluated the \textit{cultural and historic importance} as well as \textit{explicit marks} that specify the semantics of the object. 
For the structural salience, \citeauthor{Raubal2002} assessed the importance of location of the landmarks with respect to \textit{nodes}, i.e. degree of intersection and category of connecting edges, and \textit{boundaries}, i.e. objects that separate the street networks.
The main shortcoming of this work is the independence of the route; the salience of landmarks is only assessed for separate locations, i.e. intersections, and only local landmarks in the immediate surrounding are considered. 

\citet{Winter2003}  extended \citeauthor{Raubal2002}s measures with the \textit{advance visibility}, which calculates the visibility of landmarks for a person approaching a decision point.
They showed that this measure improves the suitability of the selected landmarks.
Similarly, \citet{Caduff2008} presented a framework for the assessment of landmarks salience in the visual field. 
They proposed that the landmarks salience is based on the trilateral relationship between observer, environment and geographic object. They evaluate the overall salience of visual landmarks in terms of (i) their \textit{perceptual salience}, which is based on the visual sensory input, (ii) the \textit{cognitive salience}, which is based on the prior knowledge of the individual, and (iii) the \textit{contextual salience}, which is based on the amount of attentional resources in the visual field.

\cite{Duckham2010} and \cite{Rousell2017} investigated how to automatically select the most salient landmarks from spatial databases.
\citeauthor{Duckham2010} presented an algorithm to automatically generate verbal route instructions that incorporate the most salient landmark at decision points.
Their core model is based on \textit{feature categories} of spatial databases. They assess the salience of potential landmarks by (i) evaluating the suitability of a typical instance of a POI category to be a landmark, and (ii) evaluating the likelihood that a particular instance of a POI category is typical.
In order to make the algorithm more generic, they discuss three extensions of the core model: the category weight might depend on the \textit{navigation context}, e.g. mode of travel; the \textit{side of the street} with respect to the decision points (features that lie on the same side as the turn are weighted higher); in case \textit{multiple landmarks} of the same category are in the candidate set, only the first instance of the category will get the full weight, whereas the others are dismissed.
\citeauthor{Rousell2017} presented an approach to automatically extract landmarks for pedestrian navigation from the OSM database. 
Their method is based on a number of metrics which are used to assess the overall salience of landmark candidates at decision points. 
In line with \citeauthor{Duckham2010}, they specify \textit{feature category weights} and consider the \textit{location} of the features in relation to the direction of travel (opposite side, same side). Similar to the multiple landmarks metric of \citeauthor{Duckham2010}, they specify the \textit{uniqueness} of features, i.e. if multiple feature of a category are in the candidates set, all feature get a reduced weight. Additionally they present metrics for the \textit{visibility}, the \textit{position} of the feature in relation to the decision point (before, alongside, after), and the \textit{distance}.

While the presented approaches are useful for enriching wayfinding instructions with local landmarks, the main shortcoming that was not considered over the years is the limitation to the features selection at intersections and decision points. 
It was shown that landmarks are not only important locally at decision points, but also along the route and even globally when distant to the route \citep{Anacta2017,Lovelace1999,Steck2000,Li2014a}. 
Moreover, it is criticized that current approaches focus on point-like landmarks, neglecting regional landmarks or structural regions \citep{Schwering2017,Lowen2018a}.
\cite{Sester2015} pointed out that, besides local route information, information should be provided, which embeds the route in the global context; 
\cite{Lowen2019} showed that besides landmarks, structural regions are important features in human wayfinding. 
These orientation information were shown to support the acquisition of survey knowledge.
There is a need for an algorithm to automatically select the most salient orientation information; moreover, the algorithm should not be restricted to the selection of local features at decision points, but is should automatically select orientation information for any location and context along the route.

\subsection{Orientation Information in Context}

Orientation is defined as \textit{"a dynamic process of deriving one's position in space with regard to known environmental information at a scale (or subset of scales) relevant to the current goal"} \citep{Krukar2016}.
This definition highlights the \textit{location} and the \textit{scale} as two main aspects of orientation.
In an assisted wayfinding scenario, for which we aim to automatically select orientation information, the location is not fixed, but changing with regard to the users drive along the route.
Thus, the location is one important contextual parameter.
Additionally, orientation is related to a \textit{scale relevant to the current goal}.
In previous work, a classification scheme of \textit{functional scales} in assisted wayfinding was developed, where a conceptual distinction of scales with respect to different goals in assisted wayfinding scenarios was proposed \citep{Lowen2019b}.
Maps at different scales might target different task such as supporting the identification of a decision point or providing an overview of the local or global route context.
Consequently, environmental features that are depicted in the particular maps need to be selected in order to support the task in the best possible way. 
For the scope of this paper we consider the functional scales as a second important contextual parameter, which defines the relevant size of the environment and the representation of the map.

In the following, we present an algorithm to computationally select orientation information for any point along the route. 
We extend previous approaches, i.e. we incorporate previously presented metrics, which were discussed above; we develop new metrics for the salience assessment that is not restricted to point-like landmarks and decision points only.
We parameterize the salience function to the context, i.e. the current location along the route and the target scale.

\section{Theory}
\label{sec:theory}

Landmark were shown to be important features in human wayfinding and key features in spatial cognition.
Although landmarks are defined to be any geographic object that structures human mental spatial representation \citep{Richter}, landmarks are dominantly considered as point-like objects such as a specific building.
Because of their importance to structure human mental spatial representation, landmarks are often used in human communication.
It was shown that people include local landmarks at decision points and along the route, as well as global landmarks off-route \citep{Denis1997,Daniel1998,Lovelace1999,Steck2000,Michon2001,Tom2003,Winter2008,Li2014a}.
\cite{Anacta2016} presented empirical evidence that human wayfinding instructions contain a significant amount of orientation information, i.e. information that supports people to derive their position in space and orient themselves with regard to known environmental information \citep{Krukar2016}.
We developed a classification scheme of orientation information that specifies feature types and features roles in route maps. The feature types \textit{landmarks}, \textit{network structures} and \textit{structural regions} are distinguished, moreover, the role features might take with regard to the route, i.e., local or global, is specified \cite{Lowen2019b}.

\textit{Landmarks} and can be any point-like, linear, or areal object in the environment and may be relevant in local or global context of a route.
\textit{Network structures} are defined as the relevant street network to be selected for orientation support. 
This might be on the one hand the \textit{network skeleton} constituting the overall structure of the street network (global context), and on the other hand the route relevant network including side streets and detailed network related to the route (local context).
\textit{Structural regions} comprise \textit{administrative regions} and \textit{environmental regions}, which are relevant for the global context of the route.
They where shown to support incidental spatial learning of survey information when included in route descriptions \citep{Lowen2019}.
Whereas areal landmarks are separate geographic objects with an areal extend, structural regions are in contrast defined by their \textit{bona fide} or \textit{vague} boundaries (environmental regions), or \textit{fiat} boundaries (administrative regions) \citep{Smith1998,Galton2003}, which might have containment relations with other features. 
Environmental regions have a semantic meaning, which refers to some kind of homogeneous and perceivable environmental structure, such as urban vs. rural areas or a city center.
Administrative regions might only be perceivable in the environment through signage or external reference. 
The example of a city might be considered twofold: on the one hand a city has a clearly defined administrative boundary, which is apparent via signage; on the other hand and not necessarily corresponding, a city might be considered as the build area in contrast to the surrounding rural area. 
Structural regions are rarely incorporated in current navigation systems; however, they can be useful features for supporting orientation by helping users to structure their mental spatial representations 
\citep{Lowen2019b}.

While we focused on cognitive aspects with respect to orientation support and spatial learning in previous works, in this work we focus on the computational aspects to automatically select orientation information.
Therefore, we adhere to the classification scheme of orientation information and present a selection workflow that consists of two main steps: in a first step general orientation information candidates are selected within a reasonable distance buffer around the route; in a second step the candidate set is refined with respect to the context, i.e. locational and scale context. 
Different communication modes for wayfinding support exist, however, for the proof of concepts we focus on the visual representation of orientation information on wayfinding maps.

\subsection{Selection Workflow}
\label{sec:selection_workflow}

\begin{figure}
\centering
\resizebox{\textwidth}{!}{\includegraphics{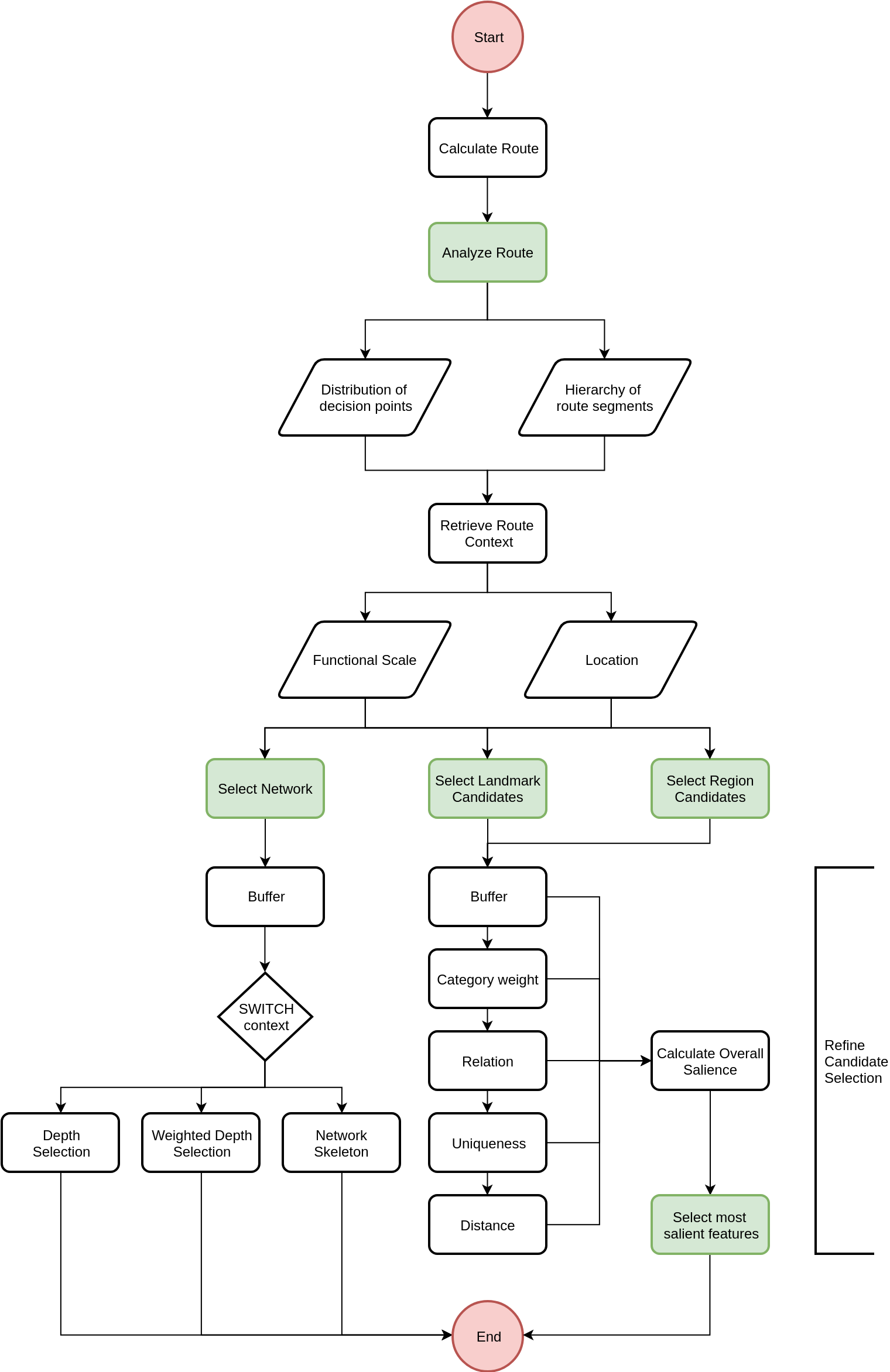}}
\caption{Flowchart of the algorithm.} \label{img:flow_chart}
\end{figure}

We implemented the following workflow in order to select orientation information for the visualization in orientation maps; we stepwise add information to the selection (see Figure~\ref{img:flow_chart}).
As described above, the relevance of environmental features for orientation support depends on the users' context in terms of the location and the scale.
We initially \textit{analyze the route} in order to better define the route context. 
We analyze the route in terms of the type of streets, the distribution of decision points, and the functional scales (Section~\ref{sec:analyze_route}). 
We then \textit{select the relevant street network} for supporting orientation. Current navigation systems only reduce the level of detail of the represented features with decreasing map scales. We expect that a selection and an accentuation of the main street network will help users to structure their mental spatial representations of space. 
Thus in Section~\ref{sec:network_structures} we elaborate a context dependent selection of the street network.
Finally, we \textit{select landmarks and structural region} to embed the route into the context. Therefore, first feature candidates are selected from spatial databases, which are then refined with respect to the salience metrics that will be described in Section~\ref{sec:select_landmarks_regions}.

\section{Algorithm}
\label{sec:algorithm}

\subsection{Route Analysis}
\label{sec:analyze_route}


The route may be computed with any existing metric ranging from classical shortest paths, such as \textit{Dijkstra} \citep{Dijkstra1959} or \textit{A*} \citep{Hart1968}, to cognitively motivated algorithms, such as \textit{simplest paths} \citep{Mark1985,Duckham2003} or \textit{fewest-turns} \citep{Jiang2011}.
The street network is organized with respect to the function and capacity of the streets; this often corresponds to the administrator of the streets. 
Specifications differ between countries and states, but generally the following distinction can be made: (i) highway; (ii) primary, secondary, and tertiary roads, which link cities, towns, and villages; (iii) residential roads which provide access to property. 
Special street types are disregarded for the purpose of the current work. 
The street types are hierarchically ordered as shown in Table~\ref{tab:street_types} (lower values represent a higher order in the hierarchy).

\begin{table}
\caption{Types of street segments.}
\begin{tabular}{rl}
\hline
Value & Type \\
\hline
10 & highway \\
20 & primary road \\
30 & secondary road \\
40 & tertiary road \\
50 & residential road \\
\hline
\end{tabular}
\label{tab:street_types}
\end{table}

The route is analyzed in terms of the hierarchy of the route segments, and the location and distribution of decision points along the route.
Routes often have a typical structure with respect to the type of route segments, i.e., increasing hierarchy at the initial part of the route and decreasing hierarchy at the final part of the route \citep{Wunderlich1982}.
The route structure is utilized for the further selection and embedding of the route in its spatial context.
Some of the above mentioned metrics consider the route structure in the route computation, e.g., natural roads \citep{Jiang2008} consider the Space Syntax theory \citep{Hillier1984}.

The street network is represented as a graph $G = (V,E)$ consisting of a set of vertices ($V$) and edges ($E$), and a route as a directed graph $G' = (V',E') \subseteq G$ from start $s \in V$ to destination $t \in V$. 
Each edge $e \in E$ is defined by two incident vertices $x,y \in V$.
Each route consists of a set of decision points $DP \subseteq V'$. 
Decision points are defined by a set $DPC$ of decision point classes (see Table~\ref{tab:dp_classes}), represented by a function $class:DP \to DPC$, such that for all $v \in DP, class(v)=c,$ with $c \in DPC$.
Street types are defined by a set $T$ (see Table~\ref{tab:street_types}), represented by a function $type:E \to T$, such that for all $e \in E, type(e)=t$, with $t \in T$.

Table~\ref{tab:dp_classes} presents the decision point classes that are used for the analysis of decision points along the route; these were presented in previous work \citep{Anacta2018}. 
\textit{Straight on} classifies decision points where the route continues straight but intersects street segments of the same or higher hierarchy class. 
\textit{Turn not at a junction} classifies only vertices $v \in V'$ with $degree(v)=2$ that require a turn, i.e., there are no side streets but the angle of connecting edges exceeds a predefined threshold. 
\textit{Highway ramp/exit} specifically classifies vertices $v \in V'$ of highways, i.e. ramps, exits, connections; thus $type(e) = 10$ with $e \in E'$ and $e$ incident to $v$.
The distribution of decision points along the route as well as the reference segments are used for further selection of environmental features. 
\textit{Reference segments} are considered as route segments before and after the particular decision point. They may either be calculated based on a predefined static distance, or a variable distance based on the decision point class, the functional scale, etc.

\begin{table}
\caption{Decision point classes.}
\begin{tabular}{rl}
\hline
DP class & DP class description \\
\hline
0 & start/destination \\
1 & straight on \\
2 & turn not at a junction \\
3 & turn at a t-junction \\
4 & turn at a junction \\
5 & turn at a roundabout \\
6 & highway ramp/exit \\
\hline
\end{tabular}
\label{tab:dp_classes}
\end{table}

\subsection{Network Structures}
\label{sec:network_structures}

Network structures are defined as the relevant street network to be selected for orientation support. 
This might be on the one hand the \textit{network skeleton} constituting the overall structure of the street network (global context); on the other hand it might be the route relevant network including side streets and detailed network related to the route (local context) \citep{Lowen2019}.
For the analysis and the selection of the relevant street network, we define the following functions:
\begin{itemize}
\item $connected(x,y)$ tests if edges $x \in E$ and $y \in E$ point to the same vertex $v \in V$.
\item $weight\_connected(x,y)$ tests if $connected(x,y)$ and $weight(x) \geq weight(y)$. 
For the $weight()$, any numeric attribute of $x$ and $y$ can be chosen; e.g. the route type (hierarchy) of the edge (see Table~\ref{tab:street_types}).
\item $depth(x)$ calculates the fewest number of connections of edge $x$ to any edge $e \in E'$ of the route.
\item $weight\_depth(x)$ calculates the fewest number of connections of edge $x$ to any edge $e \in E'$ with $weight(x) \geq weight(e)$. 
\end{itemize}
With these functions, the forthcoming operations can be performed. 
Based on the specific route context, different amounts of the adjacent route network will be considered as relevant and will be automatically selected.

\paragraph{Buffer}
The surrounding street network is reduced with respect to the users context, i.e. the network is reduced to a buffer $B$ around the current location $l \in L$ with a maximal distance $MD_f$, depending on the functional scale $f$.

\paragraph{Depth}
The depth metric selects all adjacent street segments to the route up to a pre-defined depth.
This is also referred to as \textit{topological distance} towards the route.
The set $D_n$ is the set of edges $e \in E$, for which $depth(e) = n$. 
Thus, $\sum_{i=1}^{n} D_i$ is the sum of the selected network up to the maximum depth $n$.
Depending on the functional scale, the adjacent street network is refined with respect to a specified depth.

\paragraph{Weighted Depth}
Similarly, it might be necessary to select adjacent street segments of the same or higher $weight$ with respect to the connecting route segment, i.e. weighted depth.
When driving on a primary road, side streets of a lower hierarchy would for example not be considered as relevant, whereas intersections with streets of the same or higher hierarchy are relevant to be selected.
Consequently, the set $WD_n$ represents the set of edges $e \in E$, for which $weight\_depth(e) = n$.

\paragraph{Network Skeleton}
We define the network skeleton as the overall street network of the global route context with respect to the functional scale. 
The set $NS_w$ represents the set of edges $e \in E$, for which $weight(e) \geq w$, with $w \in H$ (hierarchy classes; see Table~\ref{tab:street_types}).
Like the other metrics, parameters of this functions have to be specified with respect to specific contexts, when being implemented into actual systems. 
We assume, that at the \textit{city scale} streets of the lowest hierarchies, will not be relevant for conveying orientation, whereas at the \textit{neighborhood scale} streets of lower hierarchies are considered as part of the relevant network.
This will be evaluated in future work.

\subsection{Landmarks and Structural Regions}
\label{sec:select_landmarks_regions}

\subsubsection{Candidate Selection}
\label{sec:candidate_selection}

As mentioned above, first feature candidates for landmarks and structural regions are selected from spatial databases.
In a second step, the candidates will be refined with respect to the salience metrics that will be described in Section~\ref{sec:core_metrics}.

In spatial databases individual features are usually attributed to different feature categories, such as shops, parks, etc.
For the candidate selection, relevant feature categories need to be specified considering the data structure of the selected spatial database.
This is in line with previous research \citep[e.g.][]{Rousell2017,Duckham2010}.
All features of the specified categories that lie within a reasonable distance to the route, are considered as features candidates.
When selecting landmarks from spatial databases, the candidate set might include a large set of landmarks, thus the main requirement is to evaluate the salience of separate landmark candidates. 

The challenge for the automatic selection of environmental regions is the availability and identifiability in spatial databases, as common data structures require unambiguous feature representation. 
Although there are methods for the computational modelling of places with unclear extend \citep{Humayun2013}, environmental regions, especially vaguely defined regions, are often not available in spatial databases. Computational modelling of vague regions is mainly done for the purpose of spatial reasoning, whereas methods for the automatic detection in spatial databases do not exist.
The algorithm assumes a candidate selection based on the theoretical concept of structural regions, as it was described in Section~\ref{sec:theory}. 
In Section~\ref{sec:select_region_candidates} we demonstrate and discuss the selection of structural regions from OSM.
The candidate set will be refined with respect to the salience in a particular context, which we elaborate in the following section.

\subsubsection{Core Selection Metrics}
\label{sec:core_metrics}

In the previous section, we described the general selection of orientation information candidates from spatial databases, regardless of their relevance and salience for a specific context within a wayfinding scenario.
The selection of environmental features needs to be refined with respect to their salience for a specific context, i.e. the user's location along the route and a specific functional scale $f \in F $ at which the orientation information will be presented. 

We define the salience function $S_f$ for a weighing of the candidate set $CS$ with respect to their salience at the functional scale $f \in F$. 
In the following we describe a basic set of metrics that is used for the overall salience weighting. 
These metrics relate to the classification of \textit{visual, semantic,} and \textit{structural} salience of environmental features \citep{Sorrows1999,Raubal2002}.
The presented metrics are functions that depend on the route context, i.e. the location and the functional scale, and are applied to the candidate sets of structural regions as well as landmarks.
The metrics are not exhaustive, but additional metrics can be developed and added to the overall salience function $S_f$.
In Section~\ref{sec:extended_metrics} we discuss a few possible extensions of the salience function.

\paragraph{Buffer}
Depending on the context, only a limited area around the current location along the route will be relevant, which relates to the area of the map that will be shown on the map during wayfinding and navigation.
Thus, we refine the candidate set to a buffer around the current location $l \in L$ with a maximal distance $MD_f$. 
The buffer distance depends on the functional scale $f \in F$, which has to be specified for the specific use case.
The buffer metric is calculated as 
$$
    B_f = \left\{\begin{array}{lrr}
            1, & \text{for } & dist(l,c) \leq MD_f \\
            0, & \text{for } & dist(l,c) > MD_f
        \end{array}\right\}
$$
with $c \in CS$.

\paragraph{Category weights}
The salience of different feature categories might differ with respect to the suitability for orientation support, thus category weights might be assigned to the feature categories of the candidate set, such that $weight: C \to [0,1]$ \citep[see][]{Duckham2010,Rousell2017}. 
$C$ is a set of categories, such that for every $c \in C$, $weight(c)$ is the normalized salience of that category, with $weight(c)=1$ most salient and $weight(c)=0$ least salient.
The category weight might vary with respect to the functional scales, such that 
$$ S_{c_{f}} = weight(c_f) $$ 
is considered as the salience of any candidate of category $c \in C$ at the functional scale $f \in F$.

\paragraph{Relation}
In addition to the salience of the feature category, we specify the relation of a candidate towards the route. 
The salience of a candidate might be considered as higher when located at decision points compared to features located along the route. 
Thus, we weight candidates with respect to their relation to the route. 
We identify for every $c \in CS$ the nearest point on the route.
We consider $c$ as located \textit{at decision point}, if the nearest point on the route intersects with the reference segment of the decision point (see Section~\ref{sec:analyze_route}).
The relation metric $R$ is calculated regardless of the functional scale:
$$
    R = \left\{\begin{array}{lrr}
            1, & \text{for } & c \text{ at decision point} \\
            0.5, & \text{for } & c \text{ along the route}
        \end{array}\right\}
$$

\paragraph{Uniqueness}
If several features of the same category exist within the candidate set, the identification of a particular features in the environment is obviously more difficult. 
We therefore determine the uniqueness of the features within the candidate set \citep[see][]{Rousell2017}.
The uniqueness is calculated by 
$$
    U_c = 1/n_c
$$
where $U_c$ is the uniqueness of a feature of category $c \in C$ and $n_c$ is the number of features within the candidate set with category $c$. 
With this calculation, unique features of a category will get the value $1$, whereas multiple occurrences reduce the value (two features of the same category will both get the value 0.5).

\paragraph{Distance}
In order to distinguish local and global features we calculate the distance of the candidates towards the route.
The distance metric assigns lower values for more distant features, assuming that the salience of features decreases with increasing distance. 
Consequently, global landmarks need to be more salient with respect to the other metrics in order to get a higher overall salience, e.g. high \textit{category weight} and \textit{uniqueness}.
We define the distance metric as
$$
    D_f = 1-dist(r,c)/MD_f
$$ 
with $r = route$, $c \in CS$ and $dist$ as the euclidean distance of the candidate to the route. 
For structural regions, linear landmarks, and areal landmark, the closest point to the route on the perimeter of the features is calculated.
With $MD_f$ as the scale dependent maximum buffer distance, the distance metric decreases linear within the selected buffer.

\paragraph{Direction}
In line with considering the location context for the salience weighting, the direction of travel is specified, giving additional structure to the context.
When following a route, directions are distinguished with respect to the orientation of the user at the current location, e.g. similar to \citeauthor{Klippel2007}'s (\citeyear{Klippel2007}) turn directions. 
The relative direction of any candidate will be specified with respect to the current location and orientation as to the front, to the left or right, or to the back.  
We in general expect that features that are to the front of the current location are more relevant than features that were already driven past.
We define the distance metric as
$$
    O = \left\{\begin{array}{lrr}
            1, & \text{for } & c \text{ to the front} \\
            0.5, & \text{for } & c \text{ to the left or right} \\
            0.1, & \text{for } & c \text{ to the back}
        \end{array}\right\}
$$

\paragraph{Overall Salience}

The overall salience of the candidates with respect the functional scale $f \in F$ is calculated as
$$
    S_f = B_f * (w_s * S_{c_f} + w_r * R + w_u * U_c + w_d * D_f + w_{o} * O)
$$
with $w_s, w_r, w_u, w_d, w_o > 0$ and $\sum w = 1$.
With respect to the salience classes of \citeauthor{Sorrows1999}, we relate the \textit{uniqueness} to the visual salience, the \textit{buffer}, the \textit{distance}, the \textit{direction}, and the \textit{relation} to the structural salience, and the \textit{category weight} to the semantic salience. 
The weights $w_s, w_r, w_u, w_d, w_o$ provide the possibility to control, adjust, and optimize the influence of the particular metrics.
While the buffer metric is multiplied, thus makes a selection of features within the specified maximum distance of the current location, the other metrics are summed up, thus have a linear influence on the overall salience.  
We do not aim to provide a fixed salience function, but an algorithms that can be adjusted to the specific use case. 
Thereby, other researchers get the possibility to place weights with respect to their own interpretation or with respect to empirical findings.
In Section~\ref{sec:example} we show an example where we select orientation information candidates from OSM and apply the salience function $S_f$ to the candidate sets based on predefined weights and values for the functional scales.

\subsubsection{Extended Metrics}
\label{sec:extended_metrics}

While we presented a basic set of metrics for the salience weighting above, we want to take to opportunity to discuss a few potential extensions.
These might similarly be applied to all candidates or only be applicable to separate features types, such as structural regions or landmarks.
Additional metrics might be added to the overall salience function $S_f$ or replace previous metrics.

\paragraph{Connection}
Related to the \textit{relation} and \textit{distance} metric, the connection of candidates in the candidate set might be specified, which also relates to the structural salience.  
The connection might be distinguished in terms of the \textit{direct connection to the route}, the \textit{connection through the street network}, or \textit{no connection to the route or street network}. 
Thereby, also the distinction between local and global landmarks can be made.
Moreover, candidates can be weighted differently with respect to their connection towards the route, prioritizing directly connected features. 
In terms of the calculation, this metric would additionally consider the adjacent street network.

\paragraph{Coverage}
When considering the visual representation, especially structural regions might be too large to be visualized on the map, e.g. large environmental regions at the neighborhood scale. 
Regions that cover the whole map extract will not be identifiable from the map.
Therefore, a \textit{coverage} metric could be used, which weights the salience of regions in terms of the amount of overlap with the current map.
On the one hand, regions that cover the whole map would be disregarded; on the other hand the coverage would give higher weights to more overlap, assuming that more overlap is structurally more salient.

\paragraph{Visibility}
Previous research evaluated the visibility of landmarks at intersections \citep[e.g.][]{Raubal2002,Winter2003,Nothegger2004}.
The work might be extended with respect to the general visibility of orientation information at decision points and along the route.
While a visibility metric could easily be integrated into the previously presented salience function, we see the main challenge of this metric in the availability and processing of required data.
Moreover, the visibility might not even be a static value, but would be context dependent in many aspects, e.g. time of the day, time of the year, etc.

\paragraph{Distribution}
Especially for smaller scales that aim to provide an overview of the environment or an overview of the whole route, it might be necessary to consider the distribution of the selected features in order to avoid too much overlap and empty spots in the resulting maps.
Thus, the salience of features candidates would also depend on the location of previously selected candidates and higher weights could be assigned to features that are more distant to other features.

\section{Implementation and Evaluation}
\label{sec:example}

\begin{figure}[htb]
\centering
\resizebox*{\textwidth}{!}{\includegraphics{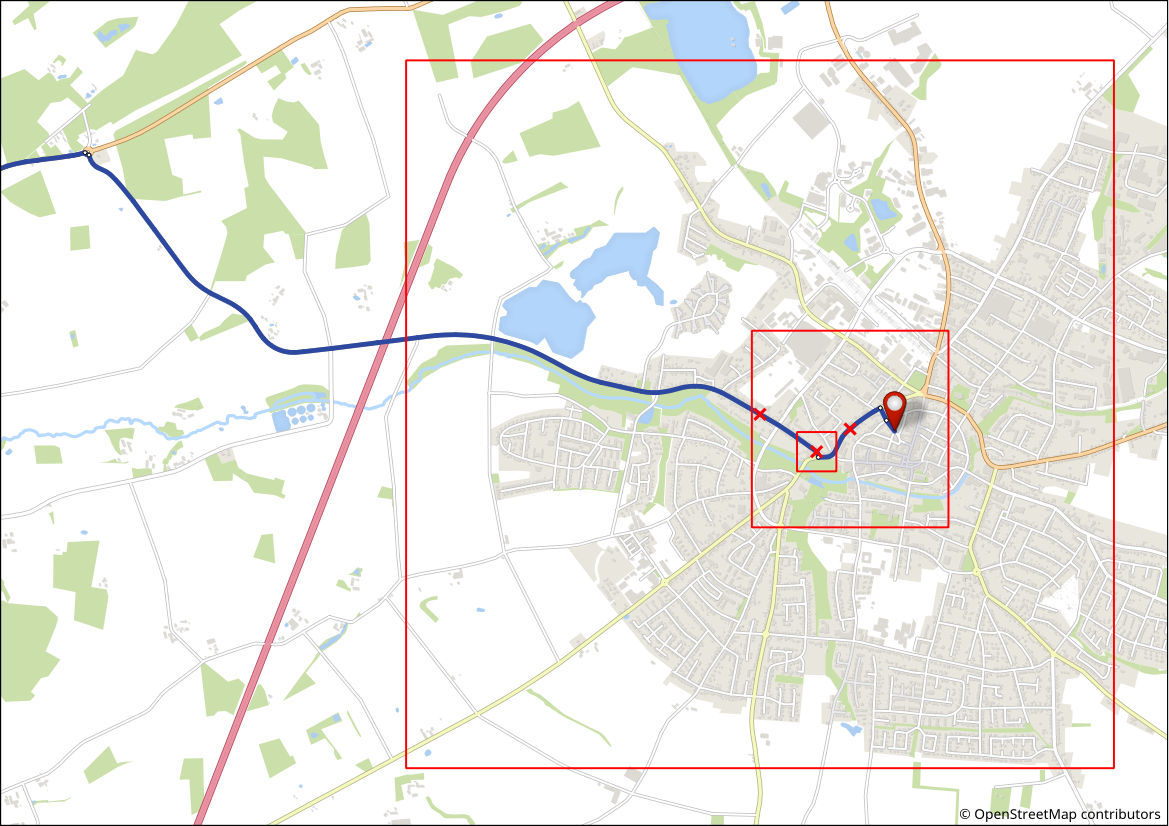}}
\caption{Selection of location (red marks) and target scales (red frames) for the selection and refinement of orientation information.} 
\label{img:route_overview}
\end{figure}

In previous work, open data from OSM was used for the selection of landmarks \citep[e.g][]{Graser2017,Rousell2017}.
In this sections, we demonstrate the previously presented selection workflow and salience weighting for orientation information using OSM as data source.
The algorithm was implemented based on a PostgreSQL database for data storage and SQL and Python for the implementation and visualization. 
We developed a QGIS plugin (\url{https://github.com/heinrichloewen/orientationMapsCreator}) which implements the functionality and can be used to calculate a route, specify the location and context for which to select the orientation information, and run the selection algorithm. We used the plugin to automatically generate the maps shown in Figures~\ref{img:route_overview} and \ref{img:example_results}.
An example route in western Germany was chosen, which covers the area of two cities of approximately 50K residents and the rural area in between; the route was analyzed as described in Section~\ref{sec:analyze_route}.
Our selection and salience weighting depends on the contextual input, i.e. a location along the route and the target scale for the map. 
As means of demonstration, we selected three exemplary locations along the route and specified the target scales (see Figure~\ref{img:route_overview}).

In the following, we describe the selection from OSM and apply the salience weighting.
Previously, methods to assign category weights based on expert ratings \citep[see][]{Duckham2010} and methods to objectively retrieve category weights, e.g. from web-harvested data \citep[see][]{Kim2016} were used.
A sophisticated category weighting is out of scope for the current work, thus we manually assign category weights based on our subjective interpretation, taking into consideration previous suggestions. 
In addition to the category weights, the overall salience functions provides parameters for weighting the different metrics. 
The weighting parameters may be used to optimize the salience weighting with respect to empirical results or expert ratings.
Different weights will influence the results of the salience function and they might even have to be optimized with respect to the influence of the metrics at different scales.
We apply equal weights for demonstration purpose. 
In Section~\ref{sec:results} and \ref{sec:discussion}, we present and discuss the results of the algorithms.
It is important to note, that the results might not be the \textit{correct} selections from the conceptual perspective, however, the aim is to demonstrate the functionality of the algorithms.
Future work needs to optimize the input parameters of the algorithms in order to generate cognitively adequate results.

\subsection{Feature selection from OSM}
\label{sec:select_region_candidates}

\begin{figure}[htb]
\centering
\includegraphics{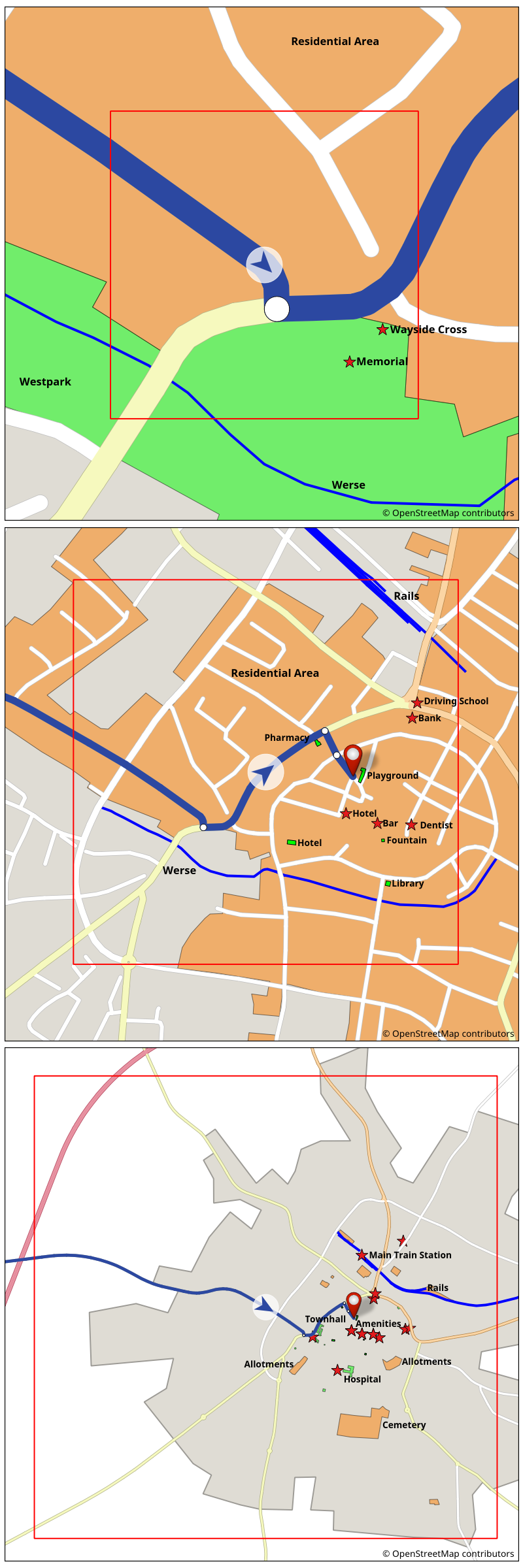}
\caption{Results for the three exemplary contexts; intersection scale (top), neighborhood scale (middle), city scale (bottom). Features are highlighted as follows: administrative regions - gray, environmental regions - orange, point landmarks - red, linear landmarks - blue, areal landmarks - green.} 
\label{img:example_results}
\end{figure}

Several researchers presented methods to select landmarks from OSM by specifying OSM tag lists \cite[e.g.][]{Rousell2017,Graser2017}. 
Although this is very selective and despite the limited availability of environmental regions in spatial databases, we use the specification in Table~\ref{tab:OSM_regions} for the candidate selection of administrative regions (AR) and environmental regions (ER) from OSM. 
Environmental regions are either defined by their \textit{boundary} tag (\textit{protected\_area, landuse, martime, national\_park}) or by their \textit{landuse} tag. 
For the former we chose the most important values that can be considered as ER from OSM taginfo (\url{https://taginfo.openstreetmap.org/keys/boundary#values}); besides, the \textit{boundary} tag contains many specific keys for different kinds of boundaries, e.g. political. 
For the latter we consider all values of the \textit{landuse} tag as potential candidates of environmental regions.
In addition, we assign category weights as shown in Table~\ref{tab:OSM_regions}.
After selecting region candidates and running the algorithm for calculating the feature salience, the results are highlighted on a map for visual interpretation (see orange regions in Figure~\ref{img:example_results}).

\begin{table}[htb]
\caption{OSM tags related to structural regions}
\resizebox{\textwidth}{!}{
{\begin{tabular}{llllll}
\toprule
\textbf{Key} & \textbf{Value} & \textbf{Requirement} & \textbf{Type} & \textbf{Description} & \textbf{Weight}\\
\midrule
boundary & administrative & admin\_level = * & AR & administrative boundary  & 1\\
\midrule
boundary & protected\_area & - & ER & boundary of & 0.8\\
 & landuse & - & ER & environmental region & 0.3\\
 & maritime & - & ER & & 0.8\\
 & national\_park & - & ER & & 0.8\\
\midrule
landuse & * & - & ER & environmental region & 1\\
\bottomrule
\end{tabular}}
}
\label{tab:OSM_regions}
\end{table}

Similarly, we specify OSM tags for the selection of landmark candidates, which are shown in Table~\ref{tab:OSM_landmarks}. 
The classification scheme we referred to in Section~\ref{sec:theory}, distinguishes point landmarks (PL), linear landmarks (LL), and areal landmark (AL). 
The OSM data structure, which consists of \textit{nodes, ways,} and \textit{relations}, however, does not correspond to the conceptual classification of different landmark types. 
Many polygonal features in OSM (closed ways) depict the boundary of buildings, thus would be processed as areal landmark candidates by the algorithm.
Thus, we consider it as important to review the data source, pre-process it if applicable, and adjust the implementation of the algorithm to the data set. 
The results shown in Figure~\ref{img:example_results} are restricted to the peculiarities of the OSM data.

\begin{table}[htb]
\caption{OSM tags related to landmarks}
\resizebox{\textwidth}{!}{
{\begin{tabular}{llp{3.5cm}lp{3.5cm}l}
\toprule
\textbf{Key} & \textbf{Value} & \textbf{Requirement} & \textbf{Type} & \textbf{Description} & \textbf{Weight}\\
\midrule
amenity     & * & name & AL,PL & amenities & 0.5\\
leisure     & * & name & AL,PL & & 0.5\\
tourism     & * & name & AL,PL & & 0.7 \\
historic    & * & name & AL,PL & & 0.8\\
shop        & * & name & PL & & 0.3\\
\midrule
barrier     & * & height OR \newline fence\_type OR \newline description  & LL & physical structure which blocks or impedes movement & 0.1 \\
\midrule
highway     & * & bridge=yes, \newline tunnel=yes & LL,PL & transport related landmarks & 0.5 \\
            & bus\_stop & - & PL & & 0.1 \\
            & crossing & - & PL & & 0.3 \\
            & rest\_area & - & PL & & 0.6\\
            & services & - & PL & & 0.4 \\
            & traffic\_signal & - & PL & & 0.3 \\
junction    & roundabout & - & PL & & 0.6\\
railway     & rail & - & LL & & 0.7\\
            & crossing & - & PL & & 0.5 \\
            & level\_crossing & - & PL & & 0.5 \\
            & platform & - & PL & & 0.5 \\
            & station & - & PL & & 0.5 \\
waterway    & * & - & LL & & 0.7 \\
\midrule
natural     & * & name & AL,LL,PL & natural landmarks & 0.3 \\
\bottomrule
\end{tabular}}
}
\label{tab:OSM_landmarks}
\end{table}

To perform the network selection (Section~\ref{sec:network_structures}), the OSM data is pre-processed in order to generate a routeable graph.
We use the osm2po tool (\url{http://osm2po.de/}) to pre-process the OSM data and save it to a PostgreSQL database. We implemented functions based on Postgis (\url{https://postgis.net/}) and pgRouting (\url{https://pgrouting.org/}) functionality to analyze the graph as described in Section~\ref{sec:analyze_route}.

\subsection{Results}
\label{sec:results}

We run the algorithm to select feature candidates and refine the selection with respect to the salience function and the specified contexts. 
In Figure~\ref{img:example_results} the results of the algorithm are shown for the selected locations and target scales.
The results are dependent on the chosen data basis and the specified weights, thus might not correspond to feature selections based on empirical investigations. 

In the top and middle map, part of the residential area is highlighted (orange polygon), which was selected as the most salient environmental region. 
However, in the bottom map, the residential area was not selected. 
When closely reviewing the separate results of the salience metrics, it turned out that the residential area got a low weight for the uniqueness metric; this is due to the OSM data structure which in this case stored the residential area as many separate small regions instead of a single large region. 
The fragmentation and granularity of OSM data depends a lot on the contributors, thus no consistency can be expected. 
Moreover, for the bottom map, many small features were selected as environmental regions (orange polygons), which would not necessarily be defined as regions. 
This is due to the availability of environmental regions in spatial databases, as it was discussed above; a more sophisticated description of potential region candidates would have to be specified when working with OSM data.
In order to prioritize larger regions that would help to better structure the environment, the \textit{coverage} metric, that was described in Section~\ref{sec:extended_metrics}, might be considered for the salience function.

For linear landmarks, a river (top and middle map) and rails (middle and bottom map) were selected. 
However, in the middle and bottom map it can be seen that only part of the features were selected. 
This again refers to underlying data structure; linear features are often divided into many features, thus the algorithm separately weights every feature.  

Furthermore, the direction metric was used to prioritise features to the front of the current location. 
The resulting maps clearly support the influence of this metric showing that predominantly features to the front of the current location are highlighted.
However, it might be argued that the applicability of this metric depends on the target scale; for the intersection scale (top map) only features to the front of the current location would be considered as relevant, whereas for the city scale (bottom map) the weighting might have to be adjusted towards a lower priority of features to the front.
Similarly, the relation metric, which prioritizes the relation towards a decision point, might be more relevant to be included in the overall salience function for larger scales, e.g. when identifying the salient features for the intersection scale.

\subsection{Discussion}
\label{sec:discussion}

While we think that the presented the algorithm can be implemented in any wayfinding support system for feature selection, few pecularities shall be discussed here, which were revealed through the exemplary implementation and demonstration: 
(i) the influence of the data basis; (ii) the influence of the metrics and metric weights; (iii) the influence of the category weights. 

The previous discussion of the results with respect to environmental regions and linear landmarks showed, that there is an influence of the \textit{data basis} on the results, namely the fragmentation of features which would subjectively be considered as a unit.
When working with OSM data, a pre-processing step is required to process the data into a consistent data format and specifically adjust the salience weighting to the underlying data basis.
Generally, the contribution of this work is in algorithm, that extends previous works on the selection and salience weighting of landmarks; we especially add functionality for the context dependence and present an adjustable and extendable method based on separate salience metrics.
When implementing the algorithm, the salience function has to be adapted to the selected data basis. 

The results are not only dependent on the underlying data basis, but also on the chosen \textit{salience metrics} and \textit{metric weights}. 
We suggested a set of salience metrics (Section~\ref{sec:core_metrics}) and discussed potential extensions (Section~\ref{sec:extended_metrics}). 
As was discussed above, these metrics need to be chosen with respect to the use case, i.e. the features types and the context.
We specified the context with respect to the functional scales, which we proposed in a separate work \citep{Lowen2019b}. 
Different contexts might serve different purposes, e.g. identifying an intersection or conveying orientation in within a city, thus different metrics and metric weights have to be specified for the salience function with respect to the current purpose.
For identifying the most salient features at a decision point, different metrics might be used than for identifying the most salient orientation information within a neighborhood or city.

Finally, for getting justifiable results, empirical evidence for the \textit{category weights} is required, which generally specifies the salience of different features types with respect to the current purpose. 
Although previous work investigated methods for specifying category weights for landmarks \citep[e.g.][]{Duckham2010,Kim2016}, the general salience of different environmental features (landmarks, network structures, structural regions) in terms of orientation support needs to be investigated in future work.

\section{Conclusions}
\label{sec:conclusions}

Previous work presented algorithms for the selection of the most salient landmarks at decision points.
However, it was shown that not only landmarks, but orientation information are important features in human spatial cognition.
In this work paper we presented an algorithm to automatically select route dependent orientation information. 
This extends previous solutions in two ways: (i) our algorithm is not restricted to the selection around decision points, but runs for any location and context along the route; (ii) beyond landmarks, our algorithms considers all kinds of orientation information.
We presented an exemplary implementation of the algorithm and discussed the results.
In future work, cognitive aspects of orientation information selection need to be investigated. This is required for the optimization of the algorithm with respect to the underlying data basis and the theoretical input. 


\section*{Funding}

This project has received funding from the European Research Council (ERC) under the European Union's Horizon 2020 research and innovation programme (grant agreement n$^\circ$ 637645).

\bibliography{references}

\end{document}